\documentclass[runningheads]{llncs}

\usepackage{url}
\usepackage{csquotes}
\usepackage{subcaption}
\usepackage[T1]{fontenc}
\usepackage{graphicx}

\begin{document}
\title{Comparative Analysis of Encoder-Based NER and Large Language Models for Skill Extraction from Russian Job Vacancies}
\titlerunning{Comparative Analysis of NER and LLMs for Skill Extraction}

\author{
Nikita Matkin\inst{1}\orcidID{0000-0003-3987-4445} \and
Aleksei Smirnov\inst{1}\orcidID{0009-0005-9684-3217} \and
Mikhail Usanin\inst{1}\orcidID{0009-0009-7733-9364} \and
Egor Ivanov\inst{1}\orcidID{0009-0007-9330-1185} \and
Kirill Sobyanin\inst{1}\orcidID{0000-0003-2224-4260} \and
Sofiia Paklina\inst{1}\orcidID{0000-0001-9666-989X} \and
Petr Parshakov\inst{1}\orcidID{0000-0002-1805-2680} 
}


%
\authorrunning{N. Matkin et al.}


\institute{HSE University, Perm, Russia \\
\email{pparshakov@hse.ru}\\
\url{https://idlab.hse.ru/en/} 
}


\maketitle

\begin{abstract}

The labor market is undergoing rapid changes, with increasing demands on job seekers and a surge in job openings. Identifying essential skills and competencies from job descriptions is challenging due to varying employer requirements and the omission of key skills. This study addresses these challenges by comparing traditional Named Entity Recognition (NER) methods based on encoders with Large Language Models (LLMs) for extracting skills from Russian job vacancies. Using a labeled dataset of 4,000 job vacancies for training and 1,472 for testing, the performance of both approaches is evaluated. Results indicate that traditional NER models, especially DeepPavlov RuBERT NER tuned, outperform LLMs across various metrics including accuracy, precision, recall, and inference time. The findings suggest that traditional NER models provide more effective and efficient solutions for skill extraction, enhancing job requirement clarity and aiding job seekers in aligning their qualifications with employer expectations. This research contributes to the field of natural language processing (NLP) and its application in the labor market, particularly in non-English contexts.

\end{abstract}
\section{Introduction}

The labor market is undergoing significant transformations driven by technological advancements and evolving economic conditions. These changes are marked by heightened demands on job seekers and a proliferation of available job openings. A major challenge in this evolving landscape is the accurate understanding of employer requirements, which vary significantly across different companies and job roles. This variability complicates the identification of essential skills and competencies necessary for job seekers to meet these requirements.

A critical issue in this context is the lack of clarity in job descriptions regarding skill requirements. Many job postings omit crucial skills, either due to oversight or an assumption that they are implicitly understood. This lack of explicit skill requirements creates barriers to professional development and upskilling, making it difficult for job seekers to effectively align their qualifications with employer expectations.

To address these challenges, this study investigates the use of advanced computational methods for automated skill extraction from job vacancies. Specifically, we compare traditional Named Entity Recognition (NER) methods based on encoders with Large Language Models (LLMs) for extracting skills from job descriptions in Russian. Our approach leverages a labeled dataset consisting of 4,000 job vacancies for training and 1,472 for testing to evaluate the performance of these methods. Besides, we collect metrics that characterize inference time and costs to compare different methods of skills extraction.

The primary objective of this research is to enhance the precision and comprehensiveness of automated skill extraction processes. By improving these processes, we aim to provide clearer insights into job requirements, thereby aiding job seekers in better understanding and meeting employer expectations. Our findings have the potential to contribute significantly to the field of natural language processing (NLP) and its application in the labor market, facilitating more efficient and effective alignment between job seekers' skills and employers' needs.

\section{Related Work} 

The task of skill extraction has gained significant interest due to the rapidly changing labor markets and the growing complexity of job descriptions. This growing interest has led to diverse approaches and methodologies aimed at accurately identifying skills from job descriptions. Researchers have explored various levels of details and nuances in skill extraction tasks. For instance, \cite{Mike2022SkillSpan} offer a detailed classification by distinguishing between skill and knowledge components. In non-English contexts, and \cite{beauchemin2022fijo} and \cite{gnehm2022fine} analyzed French and German job descriptions, respectively.

Recent research explores the use of Large Language Models for skill extraction and Named Entity Recognition tasks. LLMs show promise in handling complex skill mentions and ambiguous cases \cite{Khanh2024Rethinking}, outperforming traditional supervised models in some scenarios. However, specifically in Named Entity Recognition, LLMs still underperform compared to supervised baselines due to the structured output required by NER tasks \cite{hu2023novel,wang2023crowner}. Data augmentation techniques and embedding manipulation have been proposed to improve  skill extraction performance \cite{ul2024improving}. Studies demonstrate the effectiveness of LLMs in generating synthetic training data for skill extraction tasks \cite{Jens2023Extreme,clavie2023large}.

Novel approaches like SkillGPT utilize LLMs for efficient skill extraction and standardization \cite{li2023skillgpt}. In the broader NER context, GPT-NER transforms sequence labeling into a generation task, achieving comparable results to supervised baselines \cite{wang2023gpt}. While LLMs show potential, challenges remain. For instance, in the context of Spanish clinical NER tasks, encoder-based models still outperform generative models \cite{Guillem2024comparative,Imed2024survey}.

To sum up, related works highlight significant advancements and diverse methodologies for skill extraction in evolving labor markets. Research has detailed classifications distinguishing between skill and knowledge components in English contexts, while studies on non-English job descriptions, such as French and German, are less common. Despite the potential of LLMs in handling complex and ambiguous skill mentions, they often fall short in Named Entity Recognition tasks compared to traditional supervised models due to the structured output required by NER tasks. Models like SkillGPT and GPT-NER have made strides in efficient skill extraction, yet there remains a notable gap. Specifically, the application of LLMs and encoder-based models for skill extraction in non-English contexts, such as Russian, has not been thoroughly investigated. This study aims to fill this gap by comparing traditional NER methods with LLM approaches for extracting skills from Russian job vacancies, thus contributing valuable insights into improving automated skill extraction processes in non-English languages.

\section{Research Design}

In this study, we evaluate the performance of various methods for extracting skills from Russian job vacancies. We compare four Large Language Models and four encoder-based Named Entity Recognition models, applying the same test sample to ensure consistency. The models used are listed and discussed below.

\subsection*{Large Language Models}

\begin{itemize}
    \item GPT-4o. This model, developed by OpenAI, is a state-of-the-art model in the GPT series, known for its advanced language understanding and generation capabilities. GPT-4 excels in various natural language processing tasks, such as translation, summarization, and question-answering. Its ability to generate coherent and contextually relevant text makes it highly effective for a wide range of applications.
    \item LLAMA 3 8b. Created by Meta, LLAMA 3 represents the latest in open-source model advancements. The 9 billion parameter variant is selected for its balance between computational efficiency and high performance.
    \item GigaChat. This model, developed by Sber, is among the leading Russian language models, designed to handle the specific linguistic features of Russian language.
    \item YandexGPT. YandexGPT is optimized for Russian language tasks and is developed by Yandex. It is designed to handle the nuances of the Russian language, making it effective for translation, sentiment analysis, and content generation.
\end{itemize}

\subsection*{Fine-tuned Large Language Models}
We also test fine-tuned LLMs. We chose LLAMA 3 and YandexGPT (as Yandex provides an option for fine-tuning).
\begin{itemize}
    \item LLAMA 3 8b, adapted by Unsloth for fine-tuning with LoRA adapters. Building on Meta's latest open-source LLAMA 3 architecture, this 8-billion-parameter model offers a significant increase in capacity while maintaining efficiency. Unsloth's adaptation, combined with the use of LoRA adapters, enhances fine-tuning capabilities, making it well-suited for specialized tasks without compromising computational performance.
    \item YandexGPT, fine-tuned with the flexibility provided by Yandex's fine-tuning options, allowing for customization tailored to specific requirements.
\end{itemize}

To perform the skills extraction task using LLMs, we used the following prompt (originally in Russian, but translated to English for convenience):

\begin{displayquote}
Extract the maximum number of important skills from this job description, described in 3-4 words each without additional explanations, formulated formally, and be sure to include digital skills, if any. Do not include experience, education requirements, or working conditions. Provide a list of skills separated by semicolons. Job description text: [*job description*].
\end{displayquote}

\subsection*{Encoder-Based NER Models}

In addition to LLMs we considered four encoder-based NER models for skill extraction task, they are:

\begin{itemize}
    \item ruBERT-base-cased by DeepPavlov. It is a version of BERT tailored for the Russian language, ruBERT is fine-tuned on Russian corpora, making it effective for Russian NLP tasks.
    \item xlm-roberta-base by FacebookAI. This multilingual model is trained on diverse languages, with the base variant offering a trade-off between model size and performance for multilingual NER tasks.
    \item bert-base-multilingual-cased by Google. Supporting multiple languages, including Russian, this model is widely tested for NER tasks and known for its generalization capabilities.
    \item xlm-roberta-large-en-ru-mnli by DeepPavlov. A large version of RoBERTa fine-tuned for both English and Russian, expected to deliver high accuracy in bilingual tasks due to its size and specialized training.
\end{itemize}

All four models are all tuned with similar hyperparameters: a learning rate of 2e-5, per-device train and evaluation batch sizes of 16, and a weight decay of 0.01 during with 10 epochs. All layers except for the last were frozen.

\subsection*{Evaluation Metrics}

To assess the models and compare their performance, we used the evaluation metrics categorized into three groups: classic machine learning (ML) metrics, speed and cost. The classic ML metrics include accuracy, F1, precision, recall and ROC AUC. Speed is measured by inference time per vacancy. Costs are calculated solely for LLMs and are evaluated based on inference costs, which refer to the amount of money spent to make a request through the LLMs' API. We also take into account the size of the models. By using these metrics, we aim to identify the most effective approach for automated skill extraction in term of not only classical ML metrics but also speed and costs of inference, thereby improving the clarity of job requirements and aiding job seekers in meeting employer expectations more effectively.

Since the output of a skills extraction task is a list of skills identified from a given text, it is important to discuss how to compare the performance of different models for this task. Comparing the output of LLMs with test labels presents a challenge due to their inherent ability to alter grammatical structures and use synonyms. 
Encoder-based Named Entity Recognition models address the challenge of token classification, whereas Large Language Models are designed for token generation. LLMs possess the capability to alter the grammatical structure of final skills. For instance, the accounting skill written in Russian as "vedenie buhgalterskogo ucheta" could be represented in various forms, such as "vedenie buhgalterskogo ucheta," "vedenie buh. ucheta," "buhgalterskiy uchet," or "buh. uchet". Traditional text preprocessing techniques like stemming or lemmatization are insufficient to resolve this issue, as abbreviated word stems may differ or LLMs may substitute words with synonyms. To establish a unified metric for both encoder-based NER models and LLMs, we implemented a vector similarity-based approach.

We utilized the BGE-M3 model to compute vectors, as it demonstrates good performance in Russian language tasks, specifically semantic text similarity and paraphrase identification, according to the rating of encoders -- Encodechka\footnote{\url{https://github.com/avidale/encodechka}}. The identification of similar skills was determined using cosine similarity. Two skills were deemed equivalent if their semantic similarity was greater than or equal to 0.85. In the example of the accounting skills above, the primary skill accounting ("vedenie buhgalterskogo ucheta") was compared to other variations: "vedenie buh. ucheta" (0.92), "buhgalterskiy uchet" (0.85), and "buh. uchet" (0.72). The first two skills can be classified as close synonyms, while the last one indicates poorer similarity. This method exhibits efficacy with synonyms; for example, "verification and approval of work completion certificates" ("proverka i soglasovanie aktov vypolnennykh rabot") demonstrates a 0.85 similarity with "verification and approval of documents" ("proverka i soglasovanie dokumentov").

\section{Dataset Overview}

The train dataset consists of 4,000 job vacancies, each annotated by experts with identified skills. The data is structured with the following fields:

\begin{itemize}
    \item id: A unique identifier for each job vacancy.
    \item title: The title of the job vacancy.
    \item desc: A detailed description of the job vacancy.
    \item values: A list of dictionaries, each representing a skill mentioned in the job description. Each dictionary contains:
    \begin{itemize}
        \item start: The starting index of the skill in the job description.
        \item end: The ending index of the skill in the job description.
        \item skill: The actual skill mentioned in the job description.
    \end{itemize}
\end{itemize}

Our test sample, consisting of 1,471 vacancies, is identical for all models, ensuring a fair and consistent comparison between the traditional encoder-based NER methods and LLMs.

\section{Models comparison}

Table \ref{t1:mlmetric} presents a comparative evaluation of various models for Named Entity Recognition in terms of classical ML metrics: accuracy, F1 score, precision, recall and ROC AUC.

\begin{table}[!htbp] \centering 
  \caption{} 
  \label{t1:mlmetric} 
\begin{tabular}{@{\extracolsep{5pt}} ccccccc} 
\\[-1.8ex]\hline 
\hline \\[-1.8ex] 
 & Model & Accuracy & F1 & Precision & Recall & ROC AUC \\ 
\hline \\[-1.8ex] 
1 & DeepPavlov RuBert NER tuned & $0.73$ & $0.81$ & $0.96$ & $0.73$ & $0.54$ \\ 
2 & YandexGPT fine-tuned & $0.73$ & $0.81$ & $0.96$ & $0.73$ & $0.54$ \\ 
3 & LLAMA 3.1 fine-tuned & $0.67$ & $0.76$ & $0.95$ & $0.67$ & $0.48$ \\ 
4 & FacebookAI/xlm-roberta-base & $0.62$ & $0.72$ & $0.93$ & $0.62$ & $0.46$ \\ 
5 & google-bert/bert-base-multilingual-cased & $0.58$ & $0.68$ & $0.90$ & $0.58$ & $0.42$ \\ 
6 & gpt-4o & $0.48$ & $0.59$ & $0.90$ & $0.48$ & $0.33$ \\ 
7 & DeepPavlov/xlm-roberta-large-en-ru-mnli & $0.38$ & $0.51$ & $0.87$ & $0.38$ & $0.28$ \\ 
8 & GigaChat & $0.38$ & $0.47$ & $0.80$ & $0.38$ & $0.24$ \\ 
9 & LLAMA 3 & $0.27$ & $0.37$ & $0.73$ & $0.27$ & $0.20$ \\ 
10 & YandexGPT & $0.23$ & $0.33$ & $0.73$ & $0.23$ & $0.17$ \\
\hline \\[-1.8ex] 
\end{tabular} 
\end{table}

The highest F1 score of 0.81 was achieved by the DeepPavlov RuBert NER tuned model and the fine-tuned YandexGPT, indicating a solid balance between precision and recall. LLAMA 3.1 fine-tuned followed with an F1 score of 0.76, demonstrating its competitiveness among fine-tuned models. FacebookAI/xlm-roberta-base and google-bert/bert-base-multilingual-cased showed F1 scores of 0.72 and 0.68, respectively, highlighting their robust performance among encoder-based NER models. The gpt-4o model, representing LLMs, had an F1 score of 0.59, showing reasonable performance but not as strong as the top models. Other LLMs such as GigaChat, LLAMA 3, and YandexGPT (not fine-tuned) had lower F1 scores of 0.47, 0.37, and 0.33 respectively, suggesting reduced effectiveness for this task.

In terms of accuracy, both DeepPavlov RuBert NER tuned and YandexGPT fine-tuned achieved the highest value of 0.73. FacebookAI/xlm-roberta-base and google-bert/bert-base-multilingual-cased demonstrated relatively high accuracies of 0.62 and 0.58, respectively. Precision was highest for DeepPavlov RuBert NER tuned and YandexGPT fine-tuned at 0.96, indicating very low rates of false positives. LLAMA 3.1 fine-tuned also showed strong precision at 0.95. The gpt-4o model had a high precision of 0.90, comparable to the aforementioned models. Recall for DeepPavlov RuBert NER tuned and YandexGPT fine-tuned was 0.73, showing their effectiveness in identifying relevant entities. The recall for gpt-4o was lower at 0.48, highlighting its challenge in this area.

The highest ROC AUC of 0.54 was achieved by DeepPavlov RuBert NER tuned and YandexGPT fine-tuned, indicating a good balance between true positive and false positive rates. Lower ROC AUC scores were observed for other LLMs, with YandexGPT (not fine-tuned) having the lowest at 0.17. These results demonstrate that traditional NER models, particularly DeepPavlov RuBert NER tuned, generally outperform LLMs across most metrics. However, fine-tuned models like YandexGPT and LLAMA 3.1 show competitive performance, particularly in precision and recall, suggesting they are effective alternatives for skill extraction tasks from job vacancies in the Russian language.

\begin{table}[!htbp] \centering 
  \caption{} 
  \label{t2:speedmetrics} 
\begin{tabular}{@{\extracolsep{5pt}} ccccc} 
\\[-1.8ex]\hline 
\hline \\[-1.8ex] 
 & Model & F1 & Model size & Time per vacancy (sec) \\ 
\hline \\[-1.8ex] 
1 & DeepPavlov RuBert NER tuned & $0.81$ & 180M & 0,025 \\ 
2 & YandexGPT fine-tuned & $0.81$ & \textgreater 29B & 1,835 \\ 
3 & LLAMA 3.1 fine-tuned & $0.76$ & 8B & 10,116 \\ 
4 & FacebookAI/xlm-roberta-base & $0.72$ & 250M & 0,030 \\ 
5 & google-bert/bert-base-multilingual-cased & $0.68$ & 110M & 0,032 \\ 
6 & gpt-4o & $0.59$ & \textgreater 175B & 2,814 \\ 
7 & DeepPavlov/xlm-roberta-large-en-ru-mnli & $0.51$ & 560M & 0,071 \\ 
8 & GigaChat & $0.47$ & 29B & 2,080 \\ 
9 & LLAMA 3 & $0.37$ & 8B & 10,809 \\ 
10 & YandexGPT & $0.33$ & \textgreater 29B & 1,835 \\ 
\hline \\[-1.8ex] 
\end{tabular} 
\end{table}

The next table (Table \ref{t2:speedmetrics}) provides a comparative analysis of various models used for skill extraction from Russian job vacancies, evaluating their performance based on F1 score, model size, and inference time per vacancy. Among the models assessed, DeepPavlov RuBert NER tuned stands out for its superior performance, achieving the highest F1 score of 0.81. This model is also characterized by its relatively small size of 180 million parameters and a fast inference time of 0.025 seconds per vacancy, demonstrating both efficiency and effectiveness.

YandexGPT fine-tuned matches DeepPavlov RuBert NER tuned with an F1 score of 0.81, and despite its larger model size of over 29 billion parameters, it maintains a reasonable inference time of 1.835 seconds per vacancy. LLAMA 3.1 fine-tuned also performs well with an F1 score of 0.76, but it has a larger model size of 8 billion parameters and a longer inference time of 10.809 seconds per vacancy.

In contrast, FacebookAI/xlm-roberta-base and Google BERT/bert-base-multilingual-cased, with F1 scores of 0.72 and 0.68 respectively, offer a balanced trade-off between performance and computational demand, with moderate model sizes and slightly longer inference times.

\begin{figure}[h]
\caption{Model performance}
    \centering
    \begin{subfigure}{0.5\linewidth}
        \centering
        \includegraphics[width=\linewidth]{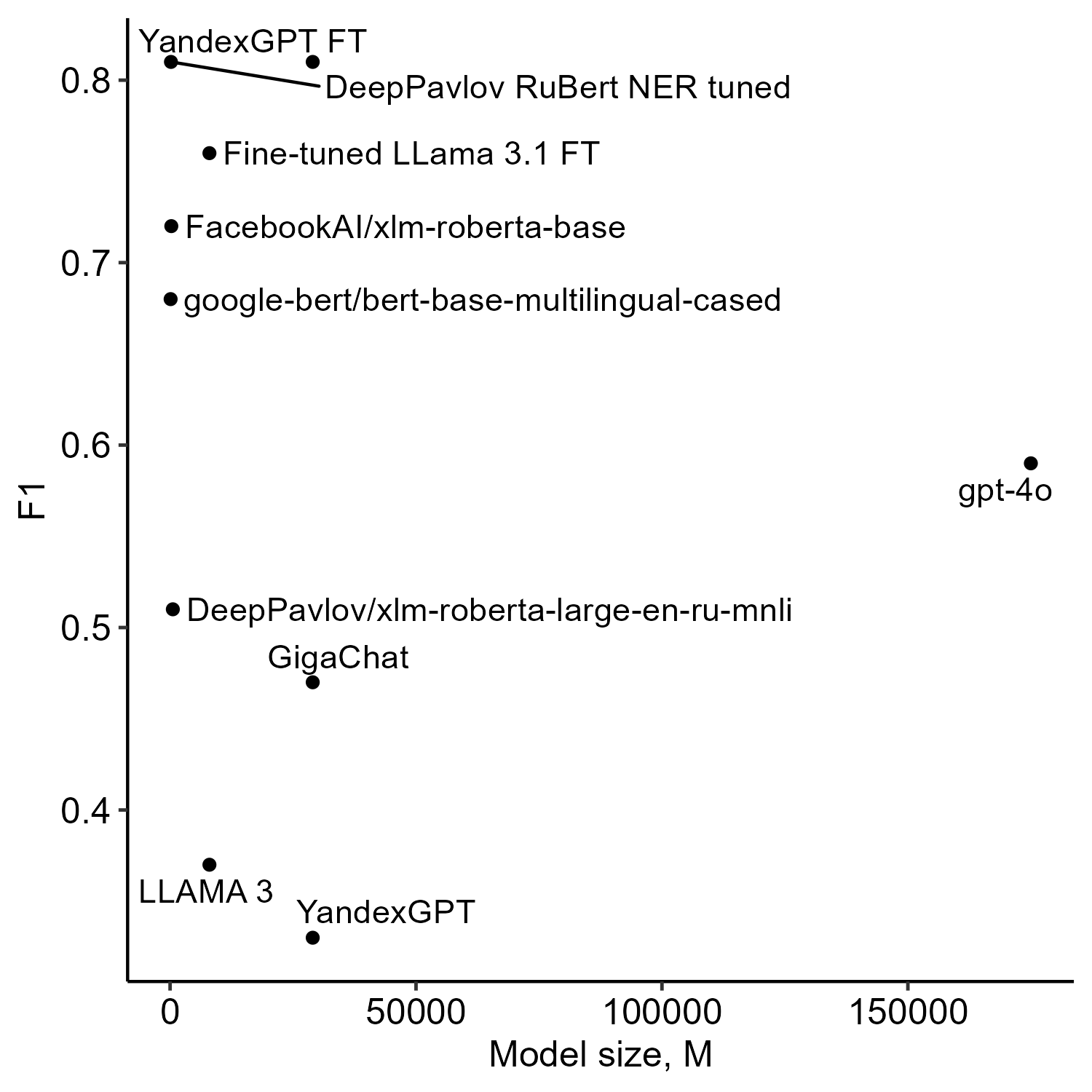}
        \caption{Model performance and size}
        \label{fig1:f1size}
    \end{subfigure}%
    \begin{subfigure}{0.5\linewidth}
        \centering
        \includegraphics[width=\linewidth]{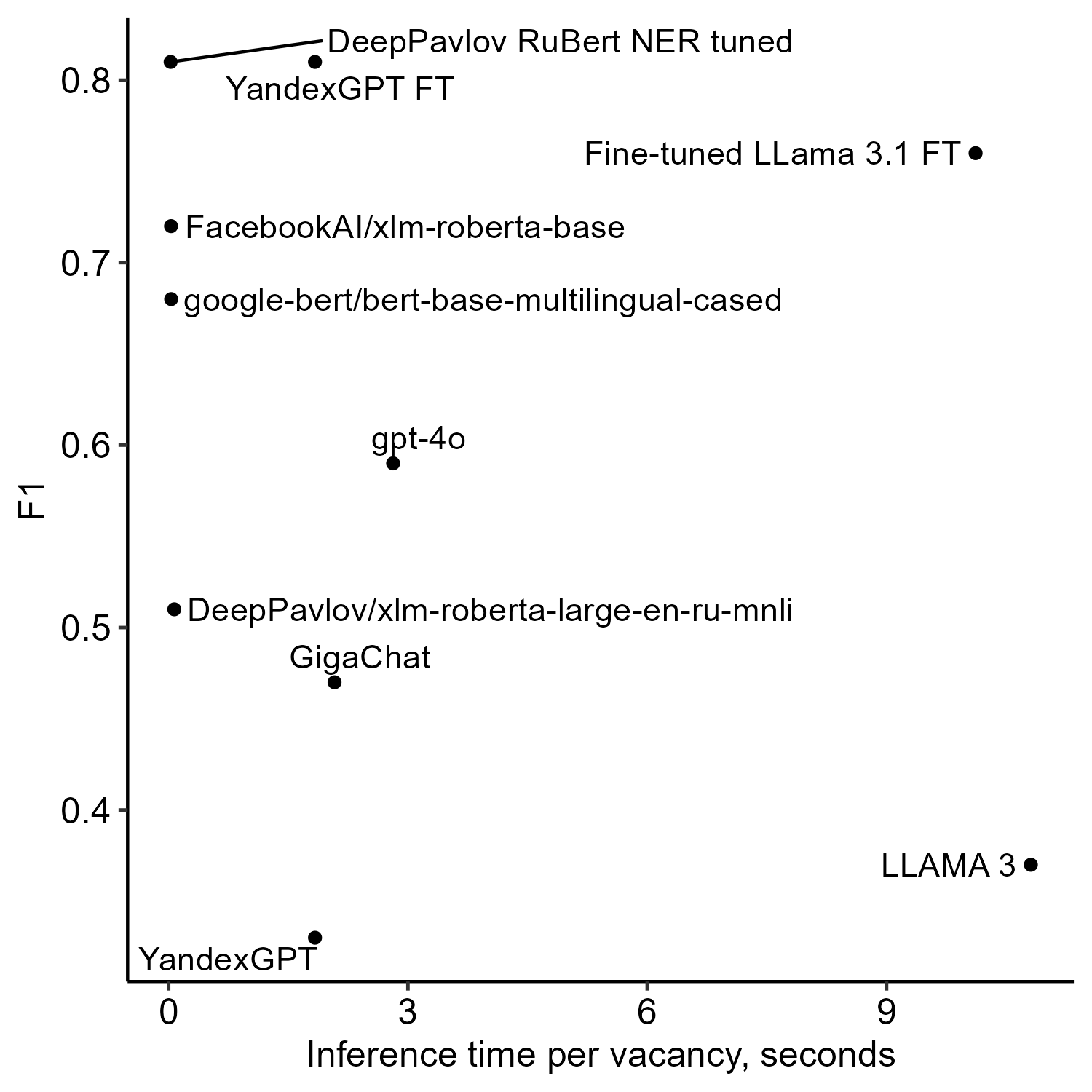}
        \caption{Model performance and inference time}
        \label{fig2:f1time}
    \end{subfigure}
\end{figure}

On the other end of the spectrum, GPT-4o, despite its massive size exceeding 175 billion parameters, shows a significant drop in F1 score and incurs a notably long inference time of 2.814 seconds per vacancy, indicating that extremely large models may not always provide proportional improvements in performance. Similarly, models like GigaChat and LLAMA 3 exhibit high inference times and substantial model sizes, with F1 scores of 0.47 and 0.37 respectively, further highlighting the limitations of large-scale models in this context. YandexGPT (not fine-tuned) presents the lowest F1 score of 0.33, with a model size exceeding 29 billion parameters and an inference time of 1.835 seconds per vacancy. Figures \ref{fig1:f1size} and \ref{fig2:f1time} illustrates these results.

Overall, traditional NER models such as DeepPavlov RuBert NER tuned outperform the evaluated large language models across the metrics, particularly in terms of F1 score and inference time, indicating their superior suitability for skill extraction tasks from job vacancies in the Russian language.

The evaluation of inference costs, conducted exclusively for LLMs, reveals significant differences in the financial efficiency of these models. The gpt-4o model incurs an inference cost of \$4.37 USD, reflecting its substantial computational demands. GigaChat demonstrates a more cost-effective profile with an inference cost of \$1.39 USD, positioning it as a more economical option among the evaluated models. Conversely, YandexGPT, with an inference cost of \$6.71 USD, is the most expensive, highlighting the high operational expenses associated with its use. These cost evaluations are crucial for determining the practicality of deploying LLMs in large-scale applications.

\section{Conclusion}

This study provides a comparative analysis of traditional Named Entity Recognition (NER) methods based on encoders and Large Language Models (LLMs) for extracting skills from Russian job vacancies. The research addresses the critical challenge of accurately identifying employer requirements, which is essential for improving job seekers' understanding of the necessary skills and competencies in the labor market and encouraging a more accurate job matching process.

Our findings demonstrate that traditional encoder-based NER models, particularly the DeepPavlov RuBERT NER tuned model, outperform LLMs across all key metrics. LLMs, despite their advanced language understanding capabilities, showed limitations in this specific application, with lower overall performance metrics and significantly higher inference times.

The study underscores the importance of selecting appropriate computational methods for skill extraction, particularly in the context of non-English languages like Russian. While LLMs offer potential benefits in handling complex linguistic structures and generating diverse outputs, traditional NER models provide more reliable and efficient solutions for structured tasks such as skill extraction.

These insights contribute to the field of natural language processing (NLP) and its application in the labor market, offering practical implications for improving automated skill extraction processes. By enhancing the precision and comprehensiveness of these processes, the research supports clearer communication of job requirements, ultimately aiding job seekers in better aligning their qualifications with employer expectations.

This study has several limitations. The dataset, though substantial, may not fully represent the diversity of Russian job descriptions across different industries and regions, limiting generalizability. The focus on Russian vacancies means results may not apply to other languages or regions without adaptation. Performance depends on training and fine-tuning processes, and further optimization could enhance results. While inference time and costs were evaluated, practical deployment considerations like infrastructure and scalability were not explored. The study did not benchmark models against human performance, and the rapid evolution of LLMs may alter future comparative performance landscapes. Addressing these limitations in future research could improve the effectiveness of automated skill extraction across various contexts.

Overall, this study highlights the continued relevance and effectiveness of traditional NER models in specific NLP tasks, while also pointing to areas where LLMs can be further optimized to meet the demands of structured information extraction in diverse linguistic contexts.

\begin{credits}
\subsubsection{\ackname} This article is an output of a research project implemented as part of the Basic Research Program at the HSE University.

\subsubsection{\discintname}
The authors have no competing interests.
\end{credits}
%
%
%
\bibliographystyle{splncs04}
\bibliography{anthology}

\end{document}